# A robust single-sensing-element tactile sensor for concurrent pressure and tackiness detection with real-time signal decoupling capability

Ying Yang[1], Mingwei Gu[1], Jia-Sen Xie[1], Xingyu Ma[1], Yan-Na Lu[1], Lin Zheng[1], Jinhui Gu[1], Junshuai Chen[1], Yunjie Lu[1], Denys Makarov[2], and Jin Ge[1*]

[1] *MOE Laboratory of Bioinorganic and Synthetic Chemistry, GBRCE for Functional Molecular Engineering, LIFM, IGCME, School of Chemistry, Sun Yat-Sen University, Guangzhou 510006, China*
[2] *Helmholtz-Zentrum Dresden-Rossendorf e.V., Institute of Ion Beam Physics and Materials Research, Dresden 01328, Germany*

[*] Correspondence author. E-mail: gejin@mail.sysu.edu.cn

## Abstract

Integrating tackiness sensation into the artificial skin of humanoid robots significantly enhances their cognitive and operational capabilities. However, existing tactile sensors face challenges in multimodal signal decoupling and stability. Here, we present a surface-soft tactile sensor that incorporates a hall sensor and a magnet within its robust elastic framework. The sensor surface indents under pressure and bulges prominently when retracted from sticky surfaces, dynamically altering the Hall sensor-magnet distance. This generates whole-process-traceable and baseline-separated signals, enabling real-time differentiation between pressure and pull-off force. This single-sensing-element design facilitates bimodal sensing at the same contact spot while eliminate stress cross-talk, enhancing both accuracy and sensitivity. The fusion of robust framework and magneto-mechanical sensing mechanism equips the sensor with exceptional reliability and excellent signal baseline stability. This tactile sensor holds substantial potential for advancing robotic capabilities in evaluating adhesive properties, monitoring rubber aging, precisely handling lightweight objects, and cognizing natural objects' surface characteristics.

# 1 Introduction

The forthcoming era of embodied intelligence urgently demands artificial e-skins that mimic or even surpass human tactile perception, endowing humanoid robots with high-level cognitive and operational capabilities.[1] Significant progresses in tactile sensors have enabled the detection of various physical stimuli, including pressure,[2,3] tensile strain,[4,5] shear force,[6,7] torsional force,[8,9] vibration,[10,11] temperature,[12,13] and humidity.[14,15] These functions allow humanoid robots to perceive physical characteristics of objects like geometry,[16] material properties,[7,17,18] textures,[19,20] and weights,[21] which are critical for dexterous object manipulation[17,22] and safe human-robot interactions.[23,24] Despite these achievements, the sensation of pull-off force, essential for evaluating surface tackiness, remains largely unaddressed. Tackiness perception plays vital role in tasks such as detecting sticky contaminants, handling lightweight objects, monitoring adhesives and coatings, assessing material aging, understanding surface properties of natural objects, and more. Its integration into humanoid robots is crucial for expanding their functionality in dynamic, real-world environments.

Accurate tackiness sensation requires simultaneous and precise measurement of both pressure and outward pulling forces at the same contact spot, as the adhesion strength is influenced by pressure magnitude, pressing duration, and retraction speed.[25] This necessitates continuous monitoring of these forces throughout the entire press-and-retract process, demanding tactile sensors capable of detecting out-of-plane bidirectional forces. Current multilayer stacked sensor structure designs,[26,27] while effective for pressure sensing, are prone to delaminate under outward pulling forces, disrupting conductive pathways or eliminating electrochemical double layers.[26,28] This makes the measurement of outward pulling forces across the entire retraction process unreliable. Efforts to mitigate this issue, such as controlled interface delamination designs,[29] often compromising sensitivity for pressure or outward pulling force. This necessitates separate senor designs for the two forces and integration of them either laterally or vertically in a single sensor platform. But this introduces further drawbacks: lateral integration fails to measure the two forces at the same contact spot, reducing the accuracy of tackiness perception, while vertical configurations suffer from stress cross-talk, compromising pressure sensitivity. Recent designs have explored alternative approaches. For instance, parallel-plate capacitive sensors with air gap channels in the capacitive dielectric layer eliminate the problem of interface delamination but have not been demonstrated for tackiness detection.[26,28] Multimodal sensors with a resistive membrane as the shared force transducer for bidirectional forces offer high sensitivity but face challenges in real-time signal differentiation due to signal overlap, hysteresis, and baseline drift.[30,31] Consequently, tactile sensors still struggle with signal instability and multimodal signal decoupling in reliable and accurate tackiness detection.

To overcome these limitations, we propose a surface-soft tactile sensor employing magneto-mechanical coupling sensing mechanism. The sensor features a robust and elastic framework integrating a Hall sensor and a magnet, serving as a shared transducer for pressure and outward pulling force. Mimicking human skin, the sensor surface deforms elastically inward under pressure and outward upon pulling force, inducing corresponding increases and decreases in the magnetic field at the Hall sensor. This generates baseline-separable signals, enabling real-time differentiation between the two forces. The sensor's ability to undergo large and elastic surface deformation ensures comprehensive monitoring of the whole detachment process. Additionally, the magneto-mechanical coupling sensing mechanism minimizes baseline instability caused by minor surface displacements, ensuring stable signal baseline. These advantages endow our sensor with exceptional reliability and accuracy to trace the entire bidirectional forces throughout the press-and-retract process. The sensor achieves a pressure sensing range of 0–150 kPa and a pulling force range of 0–33 kPa, maintaining stable sensing performance and signal baseline after 5,000 pull-off trials, 50,000 pressure cycles, and even hammer strikes. We demonstrated that a robot hand equipped with our sensor can accurately evaluate the tackiness of various surfaces and detect sticky properties of objects during manipulation tasks, highlighting its potential for advancing tactile sensing in humanoid robotics and other applications that rely on precise tackiness feedback.

# 2 Results and Discussion

### 2.1 Envisioned Applications of Tackiness Perception and Its Influencing Factors

**Figure** 1a illustrates an envisioned application scenario of a humanoid robot equipped with our tactile sensor, designed with bidirectional force sensing function for tackiness perception. This functionality allows the robot to detect sticky contamination on themselves or external objects, evaluate rubber aging degree, assess stickiness of tapes, and cognize new objects. A straightforward way for evaluating the tackiness level of a material is to keep the pressing force and contact duration constant. Materials with higher-adhesion generate a stronger pull-off force (Figures 1b–d). However, selecting suitable pressing force and contact duration is challenging due to large variation in stickiness across different materials. Low pressing force and short contact time generate negligible pull-off force for mild-adhesion surfaces, making it difficult to distinguish between low and mild adhesion. In contrast, high pressing force and long contact time result in very strong pull-off force, potentially damaging the sensor.

To design the tactile sensor with desired performance, we drew inspiration from human skin and investigated the range of tackiness it experiences in daily life. Pigskin**,** chosen for its close similarity to human skin, served as the adhesive test substrate, and adhesion tests were conducted using a standard press-and-retract process with a custom setup (Figure S1, Supporting Information). Commonly used tapes (Figure S2, Supporting Information) were selected as testing surfaces to represent the highest tackiness levels typically encountered by human skin, providing an essential reference for designing the sensor's force transducer. Results revealed that adhesion strength increased with both pressure magnitude (Figure 1e and Figure S3, Supporting Information) and pressing duration (Figure 1f and Figure S3, Supporting Information), eventually reaching saturation. These findings emphasize the importance of simultaneously detecting pressure and outward pulling forces to accurately evaluate tackiness levels. To estimate the full range of pull-off forces experienced by humans, the press-and-retract tests were performed under two extreme pressure conditions: 0.01 N with a 0-second holding time (the minimum reliable force applied by the tensile tester) and 20 N with a 15-second holding time (the maximum force exerted by the index finger[32]). The resulting adhesive strength range is shown in Figure 1g, providing valuable reference for defining the design requirements of the sensor, particularly the outward pulling force sensing element.

Surface chemical properties also influence the adhesion strength, making them a critical factor in sensor design. Therefore, we evaluated the adhesion strength and adhesion energy for several materials (Figure S4, Supporting Information) selected as potential candidates for sensor fabrication, including unmodified polydimethylsiloxane (PDMS), PDMS modified with $-NH_2$, -OH, and Gelatin on its surface (to mimic chemical groups of pigskin surface[33,34]), and thermal plastic polyurethan (TPU). Adhesion tests conducted under a compression condition of 2 N with a 0-second holding time (Figures S5 and S6, Supporting Information) showed that modified PDMS exhibited stronger adhesion strength than pigskin, likely due to reduced hydrogen bonding on pigskin's slightly greasy surface. However, the high adhesion strength of modified PDMS rises a risk of sensor framework damage, making it unsuitable for practical applications. TPU displayed adhesion force similar to pigskin, while unmodified PDMS showed weaker adhesion strength. The latter combines lower adhesion strength, which reduces the risk of framework damage, and higher resilience, which ensures sensing stability, making it more suitable choice for tactile sensor fabrication.

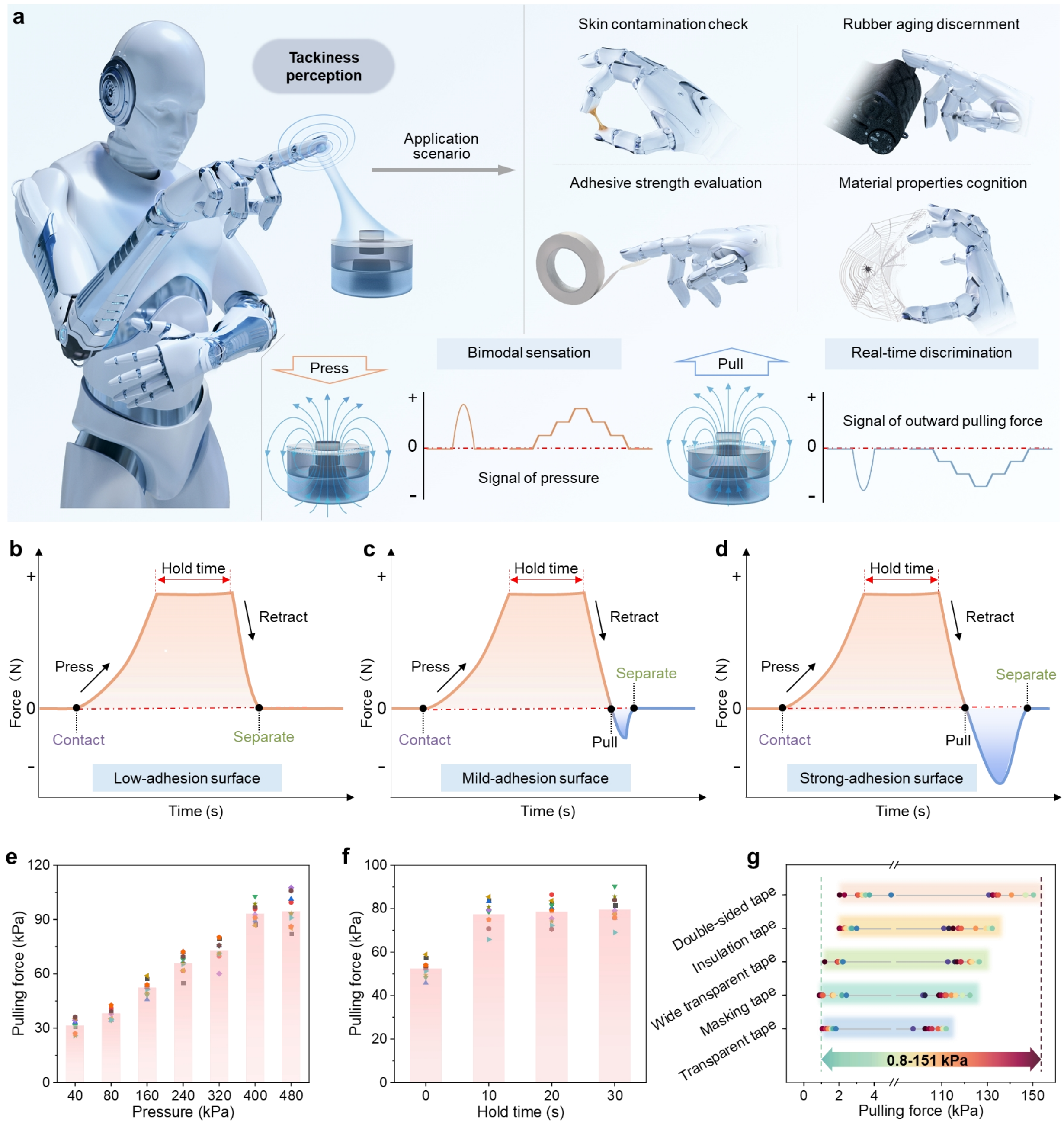

**Figure 1. Tackiness perception for humanoid robots and its influencing factors. (a)** A Robot equipped with a tactile sensor can perceive pressure and outward pulling force in real time during finger-object interactions. A conceptual diagram illustrates the diverse applications of tackiness perception, including self/foreign-object contamination detection, evaluation of rubber aging, measurement of surface adhesion properties, and cognition of new materials. **(b–d)** Schematic illustration of the real-time force experienced by human skin or synthetic skin when in contact with low-adhesion, mild-adhesion, and strong-adhesion substances. The entire process is divided into five steps: approaching, contact, maintaining force, retracting, and separate. The red dashed line above represents pressure, while the dashed line below indicates outward pulling force. Key factors influencing the adhesion strength between pigskin and double-sided tape: the magnitude of applied pressure **(e)** and the pressing duration **(f)**. **(g)** Adhesion strength range of five types of tape under two extreme pressure conditions (0.01 N with a 0-second holding time and 20 N with a 15-second holding time).

## 2.2 Design Principle of the Tactile Sensor

The deformation behavior of human skin during the press-and-retract process (Figure S7, Supporting Information) inspired the design of tactile sensor with soft surface capable of significant inward and outward deformation under pressure and pull-off force respectively (Figure S8, Supporting Information). Building on this concept, we developed a surface-soft tactile sensor utilizing magneto-mechanical coupling sensing mechanism, enable skin-like deformation for bidirectional force sensing. The structural design and working mechanism are illustrated in **Figure** 2a. It consists of a robust and elastic PDMS framework, including a ring-shaped PDMS spacer for structural support and a PDMS membrane that acts as shared force transducer for both pressure and pull-off force. A top PDMS cylinder on the PDMS membrane facilitates force transmission, while a soft magnet beneath the PDMS membrane establishes a built-in magnetic field at the Hall sensor fixed within the bottom PMMA frame. When the sensor contacts an object under applied pressure, its surface indents, reducing the distance between the magnet and the Hall sensor and thereby increasing the magnetic field at the Hall sensor. During detachment, the outward pulling force causes the sensor surface to bulge outward, increasing the distance between the magnet and the Hall sensor, decreasing the magnetic field at the Hall sensor. These opposing deformations result in corresponding increase and decrease in the Hall sensor's voltage signals. Thus the signals of pressure and pulling force are baseline-separated, with the baseline defined as the signal when the sensor surface is at its original position. This allows real-time differentiation between pressure and outward pulling force using a single sensing element, the Hall sensor.

The fabrication process of the tactile sensor is straightforward and designed for batch production using a modular preparation approach (Figures S9 and S10, Supporting Information). Figures 2b–2d showcase the elastic framework of the sensor, the fully assembled sensor, and its cross-sectional view. The magnet, a key component for the sensor, is created by incorporating neodymium-iron-boron (NdFeB) particles into PDMS (Figures S11 and S12, Supporting Information). Its softness not only allows it to integrate seamlessly into the sensor's elastic framework but also enables strong bonding to the PDMS membrane through hot pressing after oxygen plasma treatment. Additionally, the magnet demonstrated excellent magnetic field stability during compression/tension fatigue and temperature variation (Figure S13, Supporting Information), ensuring the sensor's long-term reliability and consistent sensing performance. Magnetized under a field of 29 kOe, the magnet generates a magnetic field ranging from 5 to 0.5 mT with a steep gradient extending from its central surface to a distance of 4 mm (Figure 2e and Figure S14, Supporting Information). This range provides adequate space for sensor design. In our configuration, gaps between the magnet and Hall sensor ranging from 1 to 1.8 mm were chosen to ensure high sensitivity for both pressure and pull-off force measurement.

Coupled magnetic and solid-mechanics finite element analysis (FEA) results illustrate the variations of magnetic field at the hall sensor during the press-and-retract cycle (Figure 2f). The magnetic fields under pressure and outward pulling force are distinct and non-overlapping (Figure 2g), enabling clear differentiation between the two forces. To validate the sensor's capability in detecting and decoupling pressure and outward pulling force throughout the press-and-retract process, a sticky-surfaced cuboid polymethyl methacrylate (PMMA) tip was employed for testing. As shown in Figure 2h, pressing the sensor surface caused inward deformation, increasing the sensor voltage change $\Delta V/V_0$, $\Delta V = V - V_0$, where V and $V_0$ represent the real-time and initial voltages of the Hall sensor, respectively (Figure 2i). Upon reaching a specific pressure, the tip was gradually retracted to reduce the pressure to 0. At this point, the surface returned to its original position, and the voltage change returned to the baseline value of 0. Further retraction caused the surface to bulges outward, increasing the pulling off and decreasing the voltage. The outward deformation reached up to ~300 μm before detachment, akin to human skin's response in a similar scenario. Notably, the sensor's voltage change returned to its baseline value of 0 immediately after detachment, confirming the exceptional stability of the sensor's baseline signal.

PDMS inherently suffers from viscoelasticity and hysteresis, leading to slower response speed. To address this issue, we soaked all PDMS components and soft magnets in cyclohexane prior to sensor assembly to remove unreacted monomers.[35] This treatment reduced their hysteresis as well as Young's modulus (Figures S15–S20, Supporting Information), benefiting for sensor performance improvement.

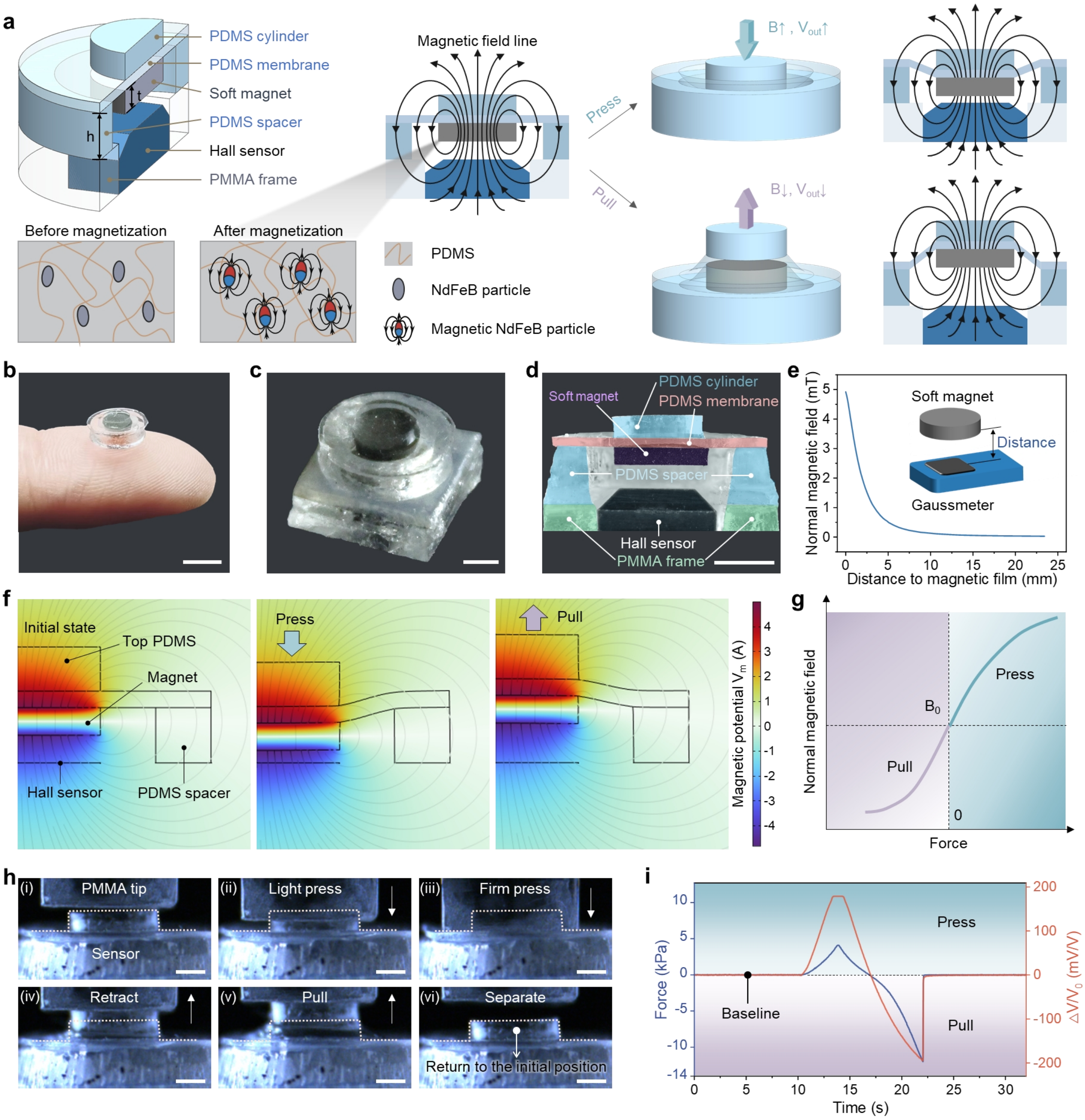

**Figure 2. Structure and working principles of the sensor. (a)** Schematic diagrams of the sensor's structure and force sensing principle. Under compressive/tensile forces, the magnetic field intensity at the Hall sensor increases/decreases accordingly. Images of the elastic framework of the sensor **(b)**, the whole sensor **(c)**, and cross-section of the sensor **(d)**, Scale bars, 5 mm, 2 mm, and 2 mm. **(e)** Normal component of the magnetic field versus distance from the center of a magnet (3 mm diameter, 0.7 mm thickness) magnetized in a 2.9 T magnetic field. **(f)** FEA results of the magnetic field distribution within the sensor in its initial, pressed, and outward-pulled states. **(g)** Example plot of Hall sensor surface normal magnetic field strength versus applied pressure and outward pulling force. **(h)** Surface deformation of the sensor during a press-and-retract cycle. **(i)** The sensor's voltage changes in response to pressure and outward pulling force.

### 2.3 Optimization of Sensor Structural Parameters

To achieve a highly compact design, the diameters of the sensor framework, internal cavity, top PDMS cylinder, and soft magnet were set to 7 mm, 5 mm, 3 mm, and 3 mm, respectively, which are the minimum dimensions required to accommodate the Hall sensor within the sensor structure. The thickness of the PDMS membrane was set to 0.3 mm for ensuring the robustness and stability of the sensor. To explore the influence of key geometric parameters—the ring-shaped PDMS membrane height $h$ and the magnet thickness $t$ (Figure 2a)—on sensing performance, a FEA model was set up to study the magnetic field distribution, stress, and deformation of the sensor. The model parameters were derived from experimental data (Figure S20, Supporting Information) and literature.[36]

FEA results reveal that the stress induced by pressure and pull-off force primarily concentrates in the region adjacent to the top PDMS cylinder/magnet, with secondary stress concentration located at the junction between PDMS membrane and the ring-shaped PDMS spacer (**Figure** 3a). Consequently, most deformation occurs in the PDMS membrane, while the top PDMS cylinder and magnet mainly undergo vertical displacement (Figure 3b). This design enhances structural sensitivity and ensures that magnetic field variations stem solely from the magnet's positional shift, avoiding complications in reconstructing magnetic stray field of shape-deformed magnet. Figures 3c and 3d illustrate that the magnetic flux density within the Hall sensor's 3 mm diameter area increases/decreases with the rise of applied pressure/outward pulling force. Influence of structural parameters, $h$ and $t$, on pressure and pulling force sensitivity are shown in Figures 3e–h. Both smaller $h$ and larger $t$ values enhances response sensitivity for the two forces.

To experimentally validate these trends, tactile sensors with various combinations of $h$ and $t$ were tested (Figure 3i and Figure S21, Supporting Information). The response signals within the range of 0–1 kPa for both pressure and pulling forces confirm that smaller $h$ and larger $t$ result in higher sensitivity (Figure 3j), closely aligning with the FEA results. However, higher sensitivity is often accompanied with reduced pressure sensing range (Figure 3k), as exemplified by the $h_{1.4}t_{0.7}$ sensor, which offers exceptional pressure sensitivity but a narrow sensing range of 0–3 kPa.

To extend the pressure sensing range without significantly compromising outward pulling force sensitivity, we filled the gap between the PDMS membrane and the Hall sensor with a PDMS sponge to increase compressive resistance (Figures S22 and S23, Supporting Information). This modification increased the saturation pressure of the Sensor_$h_{1.4}t_{0.7}$ from 1.3 kPa to 150 kPa, expending the pressure sensing range by 115 times (Figure S24, Supporting Information), albeit with a reduction in pressure sensitivity. Despite this adjustment, the outward pulling off force sensitivity remained minimally affected within the lower sensing range, achieving a balance between pressure sensing range and outward pulling force sensitivity. Upon higher pressures and faster retraction speeds, the sensor generated stronger pull-off force signals (Figures S25 and S26, Supporting Information), consistent with human's tackiness perception.

By systematically optimizing $h$, $t$, and incorporating structural enhancements like the PDMS sponge, we demonstrate how design refinements enable fine-tuning of sensitivity and sensing ranges, offering practical insights for advancing pressure and outward pulling force sensing performance. By using the optimized sensor (Sensor_$h_{1.4}t_{0.7}$_s), the entire press-and-retract process could be effectively monitored, enabling accurate tackiness evaluation. Besides the structural parameters, the sensor's response sensitivity can be adjusted by altering the Hall sensor's supply voltage (Figure S27, Supporting Information), with higher voltages offering increased sensitivity.

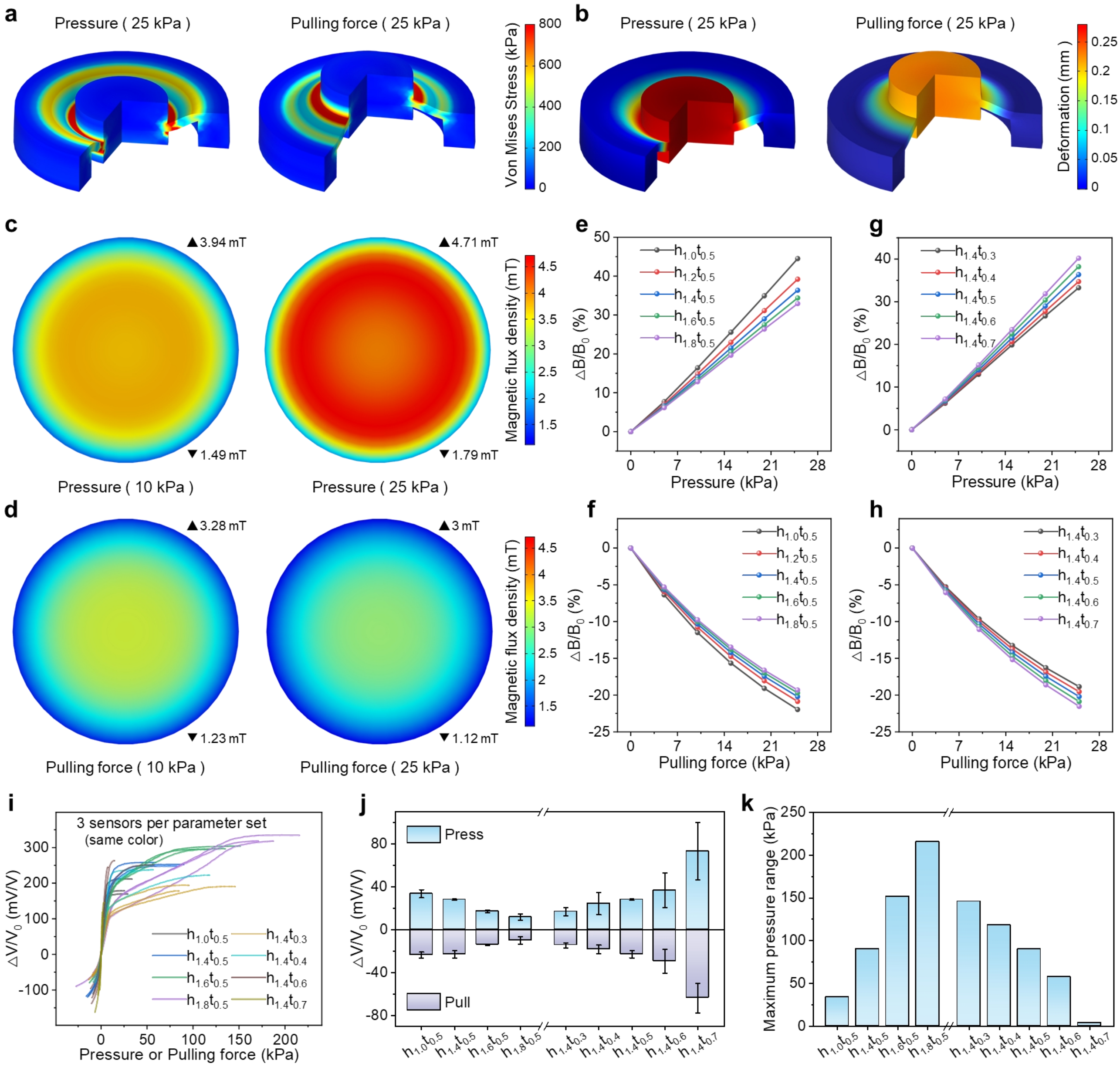

**Figure 3. Optimization of sensor structural parameters.** FEA results of the Von Mises stress distribution **(a)** and the normal deformation distribution **(b)** of the sensor under25 kPa of pressure or outward pulling force. Variation in normal magnetic flux density over a 3 mm diameter area on the Hall sensor under pressure **(c)** of 10 kPa and 25 kPa, and outward pulling force **(d)** of 10 kPa and 25 kPa. Black triangles indicate the maximum and minimum normal magnetic flux densities within the region. FEA of normal magnetic flux density changes in sensors with varying ring-shaped PDMS spacer height *h* under pressure **(e)** and outward pulling force **(f)**. FEA of normal magnetic flux density changes in sensors with different magnet thickness *t* under pressure **(g)** and outward pulling force **(h)**. The sensor with $h = x$ mm and $t = y$ mm is denoted as $h_x t_y$ or Sensor_ $h_x t_y$. **(i)** Actual performance of the sensors with different *h* and *t* combinations, with 3 sensors tested for each parameter set. **(j)** Comparison of response signal intensity for sensors with different parameter set under 1 kPa of pressure and outward pulling force. **(k)** Pressure response range of sensors with different parameter set (selecting the optimal one from three sensors).

### 2.4 Sensing Performance of the Optimized Tactile Sensor

The tactile sensor (referred to $h_{1.4}t_{0.7}$_s hereafter unless otherwise specified) were systematically evaluated in terms of reliability, bimodal decoupling ability, response speed, sensing accuracy, and stability. The schematic diagram of the testing setup is shown in **Figure** 4a. The sensor generated signals of voltage change that exhibited high consistency, with closely overlapping curves across five cycles of same press-and-retract process (Figure 4b and 4c), demonstrating its excellent reliability. Notably, the corresponding voltage changes for pressure and outward pulling force are completely separated, enabling real-time differentiation between the two forces. The sensor also exhibited rapid response and recovery times of 33.2 ms/33.5 ms for pressure and 21.8 ms/10.3 ms for outward pulling force, respectively (Figure 4d).

To verify sensing accuracy, gradient pressures were applied between the sensor and a double-side tape modified surface for the press-and-retract test (Figure 4e and Figure S28, Supporting Information). The sensor's voltage changes closely matched the applied pressure and pull-off forces (Figure 4e), highlighting its high accuracy in bidirectional force sensing. Notable, the voltage changes corresponding to outward pulling force under the same applied pressure degraded after several press-and-retract cycles (beyond 90 s, Figure 4e). This phenomenon parallels everyday experiences, where tape adhesion weakens after repeated use.

Thanks to the high sensing accuracy, our sensor can obviously reflect the differences in contact conditions and retracting speeds. Higher contact pressures, longer pressing durations, and faster retraction speeds led to more negative voltage changes at the moment of pull-off (Figures S29–S31, Supporting Information), which are consistent with results of standard tests (Figures 1e and 1f). When pure PMMA surface (low-adhesion) was used for the press-and-retract test, a counterintuitive trend was observed: higher pressures led to lower voltage changes, indicating reduced pull-off forces (Figure S32, Supporting Information). This effect likely stems from large deformation of the sensor's PDMS cylinder under high pressures, causing stress accumulation and microscale interfacial slip, ultimately reducing adhesion strength.

The sensor exhibited exceptional stability in both static and dynamic conditions. In the absence of any load, its signal baseline kept stable over 10 hours (Figure S33, Supporting Information). Under a peak pressure of approximately 63 kPa, the sensor's voltage changes corresponding to pressure and baseline remained consistent throughout 50,000 press-and-retract cycles (Figure 4f, Figure S34, Supporting Information). Fluctuations in the magnitude of the voltage changes corresponding to pull-off force were caused by the change of contact conditions (Figure S35, Supporting Information). When the contact condition was kept consistent by firmly bonding the sensor surface to the PMMA surface, the resulting voltage changes became consistent upon a same outward pulling force (Figure S36, Supporting Information).

Our tactile sensor is highly robust. Even hammer strikes caused no structural damage (Figure 4g, Movie S1, Supporting Information). The sensor's signal baseline remained stable, and it was still capable of detecting subtle pressure and outward pulling forces induced by touching the sensor surface with a finger contaminated with double-sided tape. Such excellent baseline stability is mainly attributed to the magneto-mechanical coupling sensing mechanism, which allows the PDMS membrane of the sensor to displace up to 18.2 μm without significant baseline drift (Figure S37, Supporting Information). Conventional piezoresistive and capacitive sensors, however, often experience substantial baseline fluctuations under similar displacement of their force transducers. Overall, our tactile sensor shows superior performance across multiple metrics compared with those reported in prior studies, including pressure and outward pulling force response ranges, response/recovery times, stability, repeatability, and durability (Figure 4h, Table S2, Supporting Information).

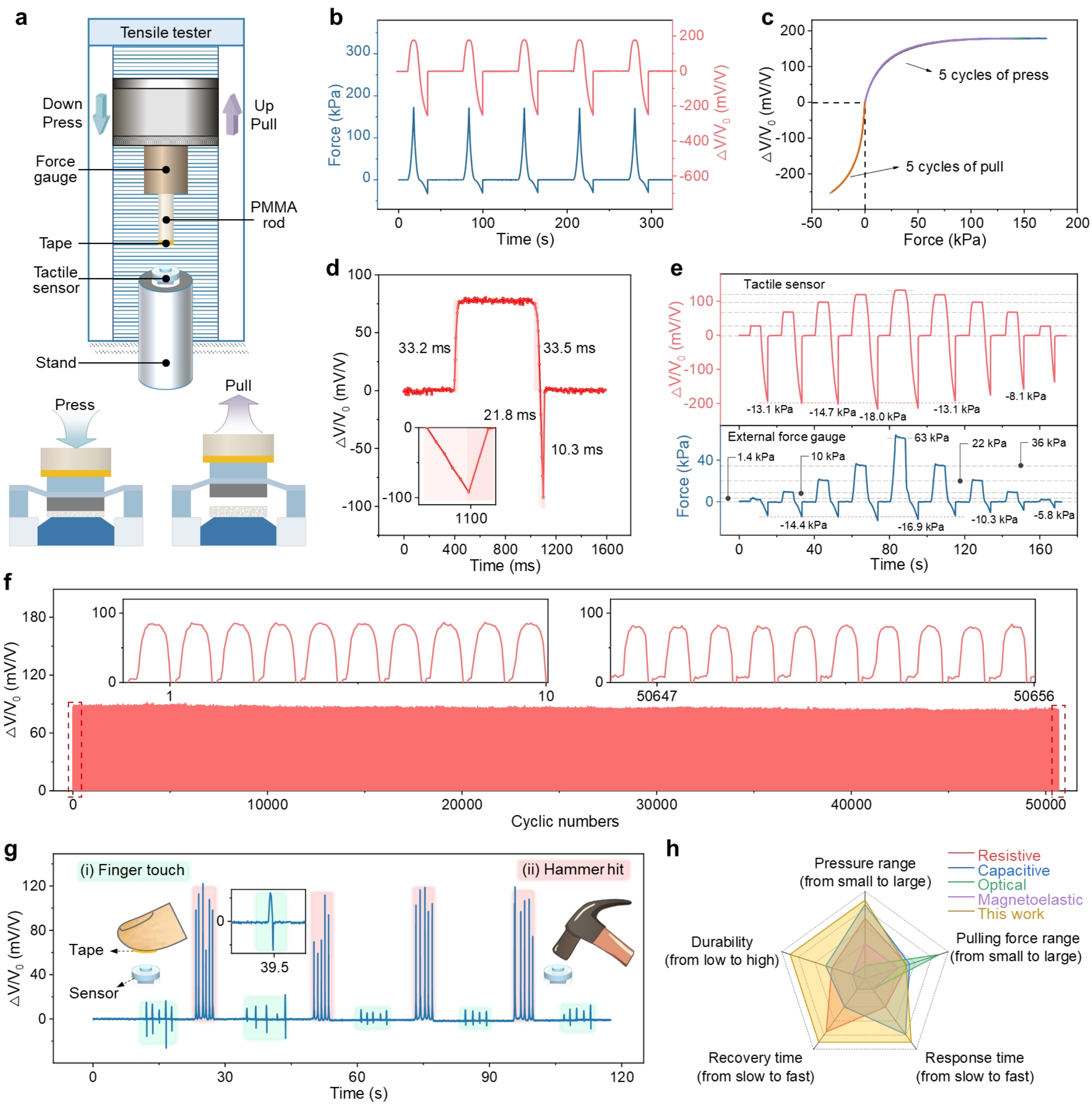

**Figure 4. Sensing performance of the sensor. (a)** Schematic of the sensor testing setup. **(b)** Sensor response to five press-and-retract cycles. **(c)** High repeatability of the response curve over five press-and-retract cycles. **(d)** Response and recovery times of the sensor under pressure and outward pulling force. **(e)** Stepwise pressure response of the sensor. The PMMA rod surface was modified with double-sided tape to apply the pulling force on the sensor. Grey dashed lines indicate consistent signal output at the same pressure, and the red dashed line shows consistent electrical signal output under a 13.1 kPa outward pulling force. **(f)** Sensor stability after more than 50,000 loading-and-unloading cycles at approximately 63 kPa pressure. **(g)** Response signal of the sensor to hammer strikes and weak pressure/pulling forces exerted by a contaminated finger. **(h)** Performance comparison of the sensor with reported sensors in terms of pressure sensing range, pulling force sensing range, response time, recovery time, and durability.

## 2.5 Applications of Tackiness Perception in Object Detection and Manipulation

While robotic hands equipped with tactile sensors have demonstrated improved performance in grasping soft and fragile objects.[17,22,37] The importance of tackiness perception remains largely overlooked. Humans may encounter trouble in handling lightweight objects with sticky contaminates on their surface. But the tackiness perception of human skin enables quick identification of this issue during object interaction, prompting humans to clean the surface before rehandling. This capability is equally important for humanoid robots.

To demonstrate it, a gripper equipped with our tactile sensor was used to grasp objects and release them afterward following the sequence shown in Figure 5a. Thanks to the high sensing accuracy and real-time bimodal decoupling capability, our sensor could monitor both pressure and outward pulling force in detail throughout the whole object handing process (Figures 5b, Figures S38–S41, Supporting Information). When the glass bottle surface was contaminated with sticky substances (simulated using double-sided tape), the robot struggled to accurately position the bottle in the targeted container. The negative voltage changes of the sensor generated in response to the outward pulling forces (Figure 5b, Movie S2, Supporting Information) alerted the robot to the need for cleaning the bottle. And after doing this, the robot could well handle the bottle, and no pull-off force signals appeared (Figure S38, Movie S3, Supporting Information). Similar results were also observed when handing a rubber bumper foot (Figures S39 and S40, Movies S4 and S5, Supporting Information). In scenarios where cleaning is inaccessible, a robotic gripper can realize dexterous object handling guided by the tactile sensor's pressure and outward pulling force sensing signals. For example, the gripper can gradually and controllably open and close to weaken the adhesion between the object and the gripper, enabling stable placement of the objects (Figure S41, Movie S6, Supporting Information).

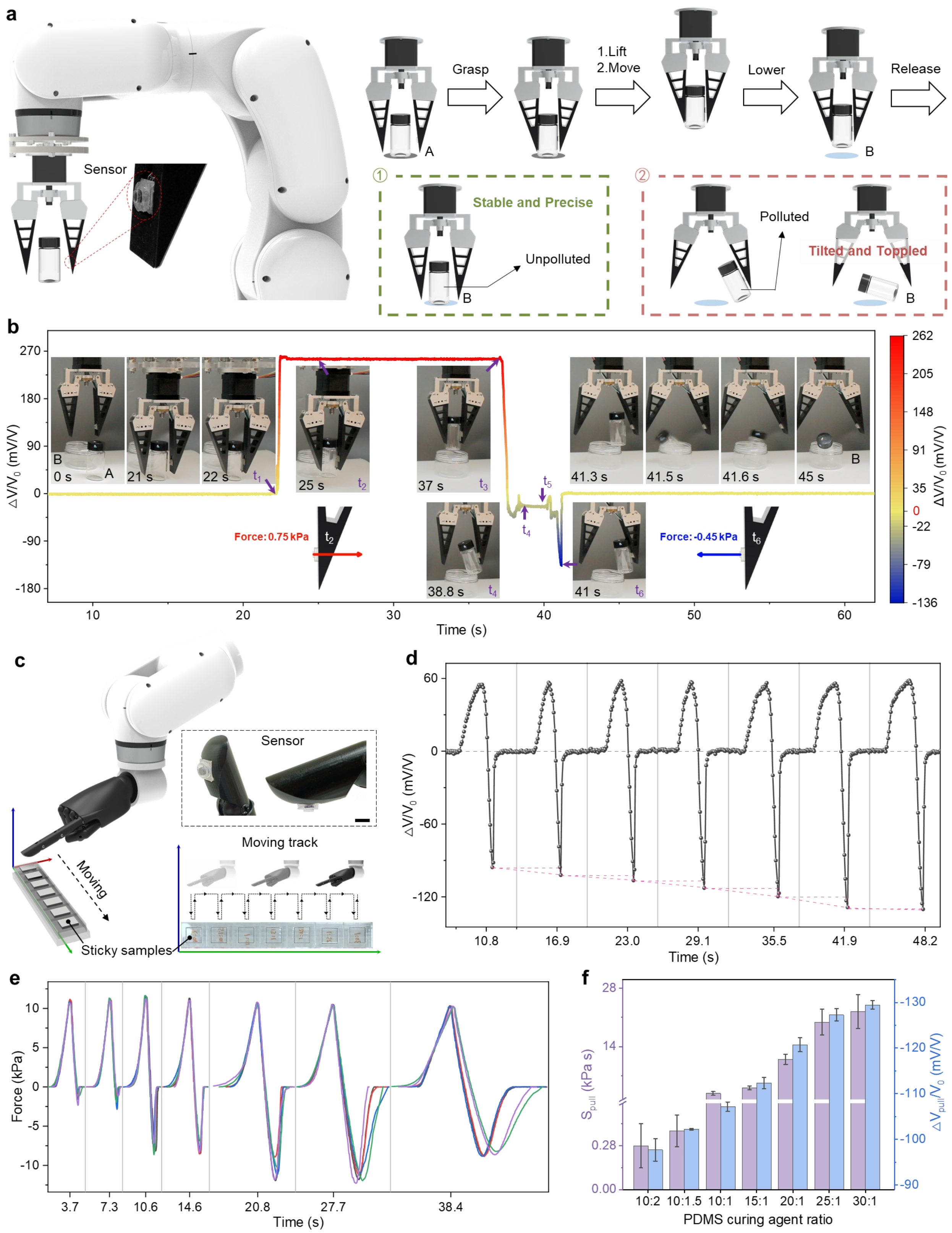

**Figure 5. Application of the tactile sensor in tackiness perception and object manipulation. (a)** Flowchart for grasping and releasing objects with a robotic arm claw equipped with the sensor. **(b)** Real-time sensor signals during the releasing process, with typical images showing the status of the robotic arm claw and object at each stage. **(c)** Schematic of a robotic hand equipped with the sensor on the fingertip to touch samples of varying tackiness. **(d)** Sensor response to samples with different tackiness levels (PDMS with base-to-crosslinker ratios of 10:2, 10:1.5, 10:1, 15:1, 20:1, 25:1, and 30:1). **(e)** Standard adhesion test of PDMS samples with different tackiness. **(f)** Comparison of tackiness ranking between standard adhesion testing (e) and sensor evaluation (d) for the same PDMS samples (mean ± standard deviation, n = 3–5).

In addition to inspect the sticky contaminations on objects' surfaces, the ability of evaluating the level of surface stickiness is also valuable for humanoid robots. Equipped with our tactile sensor, the robot's fingertip could press the objects' surfaces with same pressure magnitude and pressing duration using the pressure sensing function (Figure 5c). Then the fingertip retracted at a same fast speed, the magnitudes of voltage changes below baseline could indicate the order of adhesion strength. To demonstrate this, seven PDMS samples with varying base-to-crosslinker ratios (i.e. 10:2, 10:1.5, 10:1, 15:1, 20:1, 25:1, and 30:1) were prepared and arranged in increasing stickiness level. As the fingertip touched each sample, the pull-off force signals increased in accordance with real tackiness levels (Figure 5d, Movie S7, Supporting Information). Additional data from two repeated experiments confirm consistent results across the three trials (Figure S42, Supporting Information).

For slower retraction speed, the magnitude of pull-off force might not fully correlate with the tackiness levels (Figure 5e). Analyzing the adhesive energy may be more accurate for the assessment of stickiness. To validate this, standard adhesion tests were performed on seven PDMS samples (Figure S43, Supporting Information) at a low retraction speed using a custom setup (Figure S44, Supporting Information). Pressure and outward pulling forces were recorded using a commercial force gauge. The results showed that for base-to-crosslinker ratio below 30:1, an increase in stickiness leaded to a higher pull-off force (Figure 5e). However, the stickiest PDMS sample (30:1 ratio) exhibited a relatively lower pull-off force despite having the longest separation duration and the highest peel-off energy, likely due to its higher viscoelastic dissipation at a lower separation speed.[25] The order of adhesion energy for the PDMS samples increased with base-to-curing agent ratio (Figure 5f), indicating that peel-off energy, which encompasses the entire separation process, provides a more accurate and meaningful evaluation of stickiness than the pull-off force alone[25]. Nevertheless, at fast retraction speeds, our sensor can still evaluate the stickiness levels with accuracy comparable to the standard tests, relying solely on the magnitude for pull-off force signals (Figure 5d).

The high reliability of our tactile sensor is also worth to note. When the robot fingertip, equipped with our tactile sensor, repeatedly touched different PDMS surfaces for hundreds of cycles, the curves of voltage changes were not only nearly identical, but also accurately indicated the order of the tackiness levels (Figure S45, Supporting Information). Similar results were also observed for testing different types of tapes (Figures S46–S48, Supporting Information). These results further validate the reliability and universality of our tactile sensor in tackiness perception.

# 3 Conclusion

In summary, we developed a surface-soft tactile sensor inspired by human skin, capable of accurately and reliably detecting the tackiness of object surfaces. The robust, elastic sensor framework and the magneto-mechanical sensing mechanism are pivotal in ensuring high baseline signal stability and preserving bidirectional sensing functionality, even under significant inward and outward surface deformations. Despite its single-sensing-element design, which ensures the detection of pressure and outward pulling force at the same contact spot, the corresponding signals are baseline separated, enabling facile and real-time differentiation between the two forces. Finite element analysis and experimental results provide valuable insights for the design and optimization of the sensor. The integration of our sensor into the gripper or fingertip of robots enabled real-time detection of both pressure and outward pulling force, allowing robots to precisely place objects and differentiate surface tackiness.

While our sensor has made significant strides, there is still room for improvement, particularly in terms of the robustness and minimization. Although our sensor could withstand hammer strikes, its upper part is not yet capable of surviving large pull-off force without damage. Currently, the sensor design facilitates facile replacement of the upper part, allowing for quick and cost-effective repair when the sensor is damaged by a strong sticky surface. However, we believe that this issue could be addressed by incorporating a restraining structure that prevents the sensor surface from deforming outward beyond its maximum displacement. Regarding the minimization, we currently use the commercially available Hall sensors that have bulky encapsulation shells. Replacing these with compact Hall sensors[38–40] could significantly reduce the size of the tactile sensor, paving the way for the development of tactile sensor arrays[41,42] capable of spatio-temporal mapping of pressure and outward pulling force distributions. Overall, our tactile sensors address the challenges of reliable and precise detection of surface tackiness, enhancing the capabilities of humanoid robots in performing dexterous object manipulation tasks.

# 4 Experimental Section

**Preparation of Soft Magnet, PDMS, Modified PDMS, and PDMS Sponge**

The preparation method for soft magnet was as follows: Sylgard 184 organic silicone elastomer base, crosslinker (Dow Corning), and non-magnetized NdFeB powder (LW-BA(16-7A)-2000 mesh/5 μm; Guangzhou Xinnuo Transmission Components Co., Ltd.) were thoroughly mixed in a weight ratio of 30:3:70. The mixture was subjected to vacuum degassing at room temperature for 20 min. Subsequently, it was transferred to a clean glass slide and coated to the desired thickness using a film applicator (BEVS; 1806B; Guangzhou Mingli Instrument Equipment Co., Ltd.). The coated material was then cured on a constant temperature heating plate (GEMEI Electric; STC803-II/P2821; Shenzhen Mingrui Technology Co., Ltd.) at 90 °C for 5 min. After peeling, the material was immersed in a cyclohexane solution (Shanghai Bohr; Shanghai Bohr Chemical Reagents Co., Ltd.) for 12 h to remove any unreacted monomers, followed by cleaning with anhydrous ethanol and drying. A constant magnetic field (~2.9 T) was applied using a magnetic field control system (DXS1; Xiamen Yingdexing Magnetic Electric Technology Co., Ltd.) to permanently magnetize the soft magnet, introducing stable residual magnetization. The preparation methods for the top PDMS, PDMS membrane, and PDMS ring were similar, with a weight ratio of the base and crosslinker set at 10:1. Different tackiness PDMS samples were produced by varying the ratio of base to crosslinker (10:2, 10:1.5, 10:1, 15:1, 20:1, 25:1, 30:1) and using a 20 × 20 × 3 mm PMMA mold. The samples were then cured in an oven at 60 °C for 4 h (Shanghai Jinghong; DHG-9076Y; Shandong Zhanjia Trading Co., Ltd.). PDMS-$NH_2$ (aminated PDMS) and PDMS-OH (hydroxylated PDMS) were prepared using methods reported in the literature.[43] PDMS-Gelatin was obtained by immersing PDMS in a 5 wt% gelatin aqueous solution at a high temperature (55 °C) and then drying it.[44] The PDMS sponge was formed by soaking 1 mm thick melamine sponge in a cyclohexane-diluted PDMS precursor liquid (mixed in a 10:1 weight ratio of base and crosslinker, and combined with cyclohexane and precursor liquid in a 7:5 weight ratio). After the sponge was thoroughly saturated, excess PDMS precursor liquid was extruded, and the sponge was placed in an oven at 150 °C for 3 min to cure.

**Characterization of Soft Magnet**

The morphology of the NdFeB micro-magnets in soft magnet was characterized using a field emission scanning electron microscope (Hitachi SU8010). The elemental composition was analyzed using the attached energy dispersive spectroscopy (EDS) system. The magnetic hysteresis loop of the magnet was measured at room temperature using a Magnetic Property Measurement System (MPMS XL-7, Quantum Design) to obtain the remanence and intrinsic coercivity. The surface magnetic flux density of the magnet was detected using a handheld digital Tesla meter (DX-102F).

**Assembly of the Tactile Sensor**

Circular discs and rings of corresponding sizes were obtained by punching magnet and PDMS using hole punches of varying diameters. The magnet and top PDMS were bonded to the upper and lower sides of the PDMS membrane, respectively, through $O_2$ plasma treatment (50 W, 20 s, 100 mL/min). Subsequently, the PDMS ring was adhered to the bottom side of the PDMS membrane, ensuring that all four components were center-aligned. The resulting elastic framework of the sensor was attached to a silicone rubber adhesive (Kraft K-907) on a laser-cut PMMA support plate with a specific shape. The Hall sensor (YS1493-D) was fixed onto the laser-cut PMMA plate using 707 glue, and the PDMS sponge was subsequently attached to the Hall sensor with glue. Finally, the elastic framework of the sensor was bonded to the PMMA plate securing the Hall sensor, and the assembly was left at room temperature for 24 h to achieve optimal bonding strength. Optical images were captured using a Canon camera (EOS 4000D) or an upright metallurgical microscope (AOSVI; M203-HD228S; Guangzhou Jingchuang Computer Co., Ltd.).

**Standard Adhesion Test**

The adhesion strength between pigskin and various adhesive tapes was determined according to the standard peel adhesion methods (ASTM F2258-05 or ASTM D2979). The pigskin was cut into round pieces of 4 mm in diameter with a hole punch. Excess fat from the underside of the pigskin was carefully removed with a knife. The prepared pigskin discs were then adhered to an PMMA indenter using electrical welding glue. Finally, the assembly was mounted onto the fixture of the tensile tester. Peel adhesion tests were conducted using a tensile tester (10 N/500 N force sensor, EM2.502; Shenzhen Tesmat Instrument Co., Ltd.). The pigskin attached to the force sensor was brought into contact with the test sample (with the adhesive side facing up and the opposite side secured to the support) and then separated. The load-displacement adhesion curves were recorded during the test. All tests were performed at a constant speed of 5 mm/min, with variables including the test samples, applied force, and the pressing duration. The adhesion strength was determined by dividing the maximum peel force (peak tensile force) by the adhesive area, while the adhesion energy was calculated as the integral area under the peel force versus separation distance curve. The adhesion strength of C-PDMS (PDMS cleaned by soaking in cyclohexane) to PDMS of varying tackiness was also measured using the same methods and equipment. The C-PDMS was cut into 1 × 1 cm squares and bonded to the PMMA indenter using Kraft silicone rubber adhesive K-907. The test samples were fixed to a support with glue. The contact and separation processes were performed at constant speeds of 5 mm/min and 10 mm/min, respectively, with a 0-second holding time.

**FEA Simulation of Sensing Unit**

A two-dimensional axisymmetric model of the simplified sensor unit was established, utilizing COMSOL Multiphysics for coupled multiphysics simulations of solid mechanics and magnetic fields without current. The analysis focused on the von Mises stress distribution and the variation in normal magnetic field strength at the base during uniaxial compression and tension processes. In the solid mechanics setup, the base was defined as a fixed constraint, while the top surface was subjected to boundary loads representing applied normal pressure or tension. All materials were modeled as linearly elastic, with material density and Young's modulus obtained from experimental measurements. The Poisson's ratio values were sourced from literature,[36] and specific numerical data can be found in Table S1 (Supporting Information). For the magnetic field analysis under no current conditions, the soft magnet was designated as a permanent magnet, with a residual magnetic flux density along the z-axis set to 30 mT. The relative permeability for all materials was assigned a value of 1.

**Mechanical and Sensing Performance Test**

Compression and tension tests were conducted using a tensile tester equipped with a 10 N force sensor (EM2.502). The samples for tensile testing were laser-cut into specified shapes following the ISO 527-2 standard, while the samples for compression testing were fabricated into cylindrical shapes with a diameter of 7 mm and a height of 1.4 mm. The Young's modulus was determined through linear fitting of the obtained stress-strain curves. During the sensor performance testing, an PMMA plate affixed with double-sided tape was used as the contact indenter and secured in the tensile tester. The sensor was supplied with a 5 V supply voltage (6 V during the response time test) provided by a digital source meter (Model 2401, Keithley, Tektronix Co., US). The output voltage signal was measured using a digital multimeter (DMM7510 7½, Keithley, Tektronix Co., US), while video recordings were captured by a high-speed industrial camera (MS03130). The pressure cycling stability test was conducted using a self-assembled device controlled by a servo motor (PFDE, DB80-02430A6-A). In this setup, the sensor was mounted on a movable slider driven by a ball screw, while the PMMA indenter remained fixed. The servo motor facilitated precise movement of the sensor during the test. The tension cycling stability test was conducted using a tensile tester equipped with a 10 N force sensor.

**Robot Arm Grasping and Touch Operation**

The commercial robotic arm (Anuo Robotics; SJ603-A; Ant Automation Equipment (Shenzhen) Co., Ltd.) was programmed to execute a series of predefined motions as outlined in the experimental protocol. For the grasping tasks, the sensor was securely fixed within a 3D-printed claw groove (Creality 3D; HALOT-SKY) and mounted onto a two-finger electric claw (WHEELTEC, FAE2086), which was integrated with the robotic arm. This setup was utilized for grasping objects that were either clean or contaminated with double-sided tape. The extension and retraction of the claw were remotely controlled by a stepper motor controller (YF-18). For the tackiness differentiation task, the sensor was fixed in a 3D-printed recess on the index finger, allowing it to be integrated with the robotic hand. In both tasks, a National Instruments data acquisition card (USB-6211) supplied the sensor with a 2 V voltage, acquired the sensor's output electrical signals, and transmitted them in real time to a LabVIEW interface on the laptop.

**Conflict of interest**

The authors declare no conflict of interest.

**Acknowledgments**

This work was supported by National Natural Science Foundation of China (22475242), the Guangdong Basic and Applied Basic Research Foundation (2025A1515010271), and was supported in part via the ERC (European Research Council) grant 3DmultiFerro (101141331) and European Commission (project REGO: Cognitive robotic tools for human-centered small-scale multi-robot operations, 101070066).

**Author contributions**

Jin Ge conceived the project idea. Ying Yang and Jin Ge designed the experiments. Ying Yang performed simulations and conducted majority of experiments. Jin Ge directed the study and designed the visual interface for demonstrations. Mingwei Gu assisted in interface design and the construction of experimental setups. Jia-Sen Xie assisted with sensor fabrication and other experimental setups. Xingyu Ma implemented the machine learning algorithm and aided in testing. Yan-Na Lu provided valuable advice and assistance with the experiments and manuscript preparation. Denys Makarov provided valuable advice and assistance with experiments, analysis and interpretation of the acquired data as well as manuscript preparation. Lin Zheng assisted with the standard adhesion test and image recording. Jinhui Gu helped with performance testing and image recording. Junshuai Chen carried out the hysteresis loop characterization. Yunjie Lu assisted with scanning electron microscopy and energy dispersive spectroscopy imaging. Ying Yang and Jin Ge wrote the paper with the contribution from Denys Makarov. All authors approved the final version of the manuscript.

**Supplementary material**

Supplementary data to this article can be found online.